\def\year{2022}\relax
\documentclass[letterpaper]{article}
\usepackage{aaai22} 
\usepackage{times} 
\usepackage{helvet} 
\usepackage{courier} 
\usepackage[hyphens]{url} 
\usepackage{graphicx} 
\usepackage{xcolor}
\usepackage{graphicx} 
\usepackage{natbib} 
\usepackage{caption} 
\DeclareCaptionStyle{ruled}%
{labelfont=normalfont,labelsep=colon,strut=off}
\usepackage{booktabs}
\usepackage{algorithm}
\usepackage{algorithmic}
\usepackage{newfloat}
\usepackage{listings}
\floatstyle{ruled}
\newfloat{listing}{tb}{lst}{}
\floatname{listing}{Listing}

\usepackage{multirow}

\graphicspath{ {./images/} }

\title{Deeper Clinical Document Understanding\\ Using Relation Extraction}
\author{
    Hasham Ul Haq,
    Veysel Kocaman,
    David Talby
}
\affiliations{
    John Snow Labs Inc.\\
    
    16192 Coastal Highway\\
    Lewes, DE , USA 19958\\
    \{hasham, veysel, david\} @ johnsnowlabs.com\\
}

\begin{document}
\maketitle

\section{Abstract}



The surging amount of biomedical literature \& digital clinical records presents a growing need for text mining techniques that can not only identify but also semantically relate entities in unstructured data. In this paper we propose a text mining framework comprising of Named Entity Recognition (NER) and Relation Extraction (RE) models, which expands on previous work in three main ways. First, we introduce two new RE model architectures – an accuracy-optimized one based on BioBERT and a speed-optimized one utilizing crafted features over a Fully Connected Neural Network (FCNN). Second, we evaluate both models on public benchmark datasets and obtain new state-of-the-art F1 scores on the 2012 i2b2 Clinical Temporal Relations challenge (F1 of 73.6, +1.2\% over the previous SOTA), the 2010 i2b2 Clinical Relations challenge (F1 of 69.1, +1.2\%), the 2019 Phenotype-Gene Relations dataset (F1 of 87.9, +8.5\%), the 2012 Adverse Drug Events Drug-Reaction dataset (F1 of 90.0, +6.3\%), and the 2018 n2c2 Posology Relations dataset (F1 of 96.7, +0.6\%). Third, we show two practical applications of this framework – for building a biomedical knowledge graph and for improving the accuracy of mapping entities to clinical codes. The system is built using the Spark NLP library which provides a production-grade, natively scalable, hardware-optimized, trainable \& tunable NLP framework.

\section{Introduction}

Biomedical literature has witnessed exponential rise in the past decade. MEDLINE currently holds more than 26 million records from 5639 publications, and has indexed more than 5 million records in the past seven years alone \cite{DBLP:journals/corr/abs-2009-09509}. Furthermore, public databases like \textit{https://clinicaltrials.gov} have seen an explosion of trials data as the aftermath of the novel Covid-19 outbreak.

In addition, wide-spread adoption of Electronic Health Records (EHRs), has made copious amount of free-text data available in digital format. This unstructured data is usually documented by healthcare professionals during the course of patient care, such as clinical notes, discharge summaries, lab reports, and pathology reports \cite{wei2019relation}. While publications and literature are growing rapidly, there still lacks structured knowledge that can be easily processed by computer programs. Relation Extraction becomes even more pertinent in biomedical research as it can provide the critical links required to generate knowledge graphs for better analysis and research, and even text summarization. Relating entities also help us improve medical coding by enriching vanilla entity chunks with surrounding information to get more accurate codes.



\begin{figure}[h!]
  \includegraphics[width=0.45\textwidth]{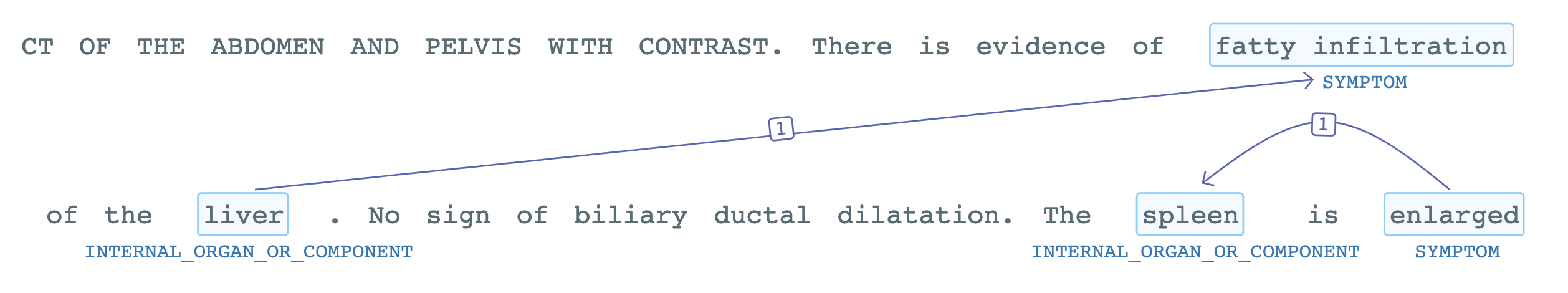}
  \caption{A Relation Extraction model semantically relating symptoms and body organs in a sample text.}
  \label{fig:re_example_1}
\end{figure}

Relation extraction is generally regarded as a classification problem where entity pairs - usually identified by NER models - are classified to determine their relationship type in a given context. These models are trained to identify semantic relations between recognized entities as illustrated in Figure \ref{fig:re_example_1}.

Since the classification implicitly relies on context, transformers based models like BERT \cite{DBLP:journals/corr/abs-1810-04805} have been shown to outperform traditional methods of dependency parsing. Recently, there is also an increasing trend of jointly training large BERT models on NER and RE tasks with shared layers and features \cite{DBLP:journals/corr/abs-2010-03851}. However, even in joint learning, the RE classification is still contingent upon entity spans identified by the NER model.

While the trend of training large transformer models continues, applying them on large datasets remains a challenge as they require significant computational resources. Furthermore, long documents containing high number of entity spans can exponentially increase probable entity pairs for RE classification - requiring significantly more resources and processing time.

In this study we focus on three major aspects of RE; the model architectures and their scalability, evaluating the models on benchmark datasets, and training and using RE for general use-cases. We also study the application of RE for understanding different aspects of clinical documents like extracting and relating dates to generate timeline of a patient's data on a timeline, or parsing and understanding trial results on large cohorts for analysis. 

Following are the novel contributions of this paper:
\begin{itemize}
\item Introducing two new RE architectures.
\item Evaluating and comparing performance of the proposed models on benchmark datasets.
\item Training the models on custom datasets and demonstrating how RE can be used to get a structured output for specific use-cases.
\item Studying the use-case of putting the history and medical history of patients on a timeline.
\item Analyzing the benefits of using RE to get more precise entity chunks for achieving better performance while mapping them to medical codes.
\end{itemize}

\section{Approach}

We treat RE as a classification problem where each example is a pair of biomedical entities appearing in a given context - the entities being NER chunks, and context being the sentence / entire document - and develop two novel solutions; the first one comprising of a simpler FCNN architecture for speed, and the second one based on the BioBERT \cite{DBLP:journals/corr/abs-1901-08746} architecture for accuracy. We experiment both approaches and compare their results.

\begin{figure}[h!]
  \includegraphics[width=0.45\textwidth]{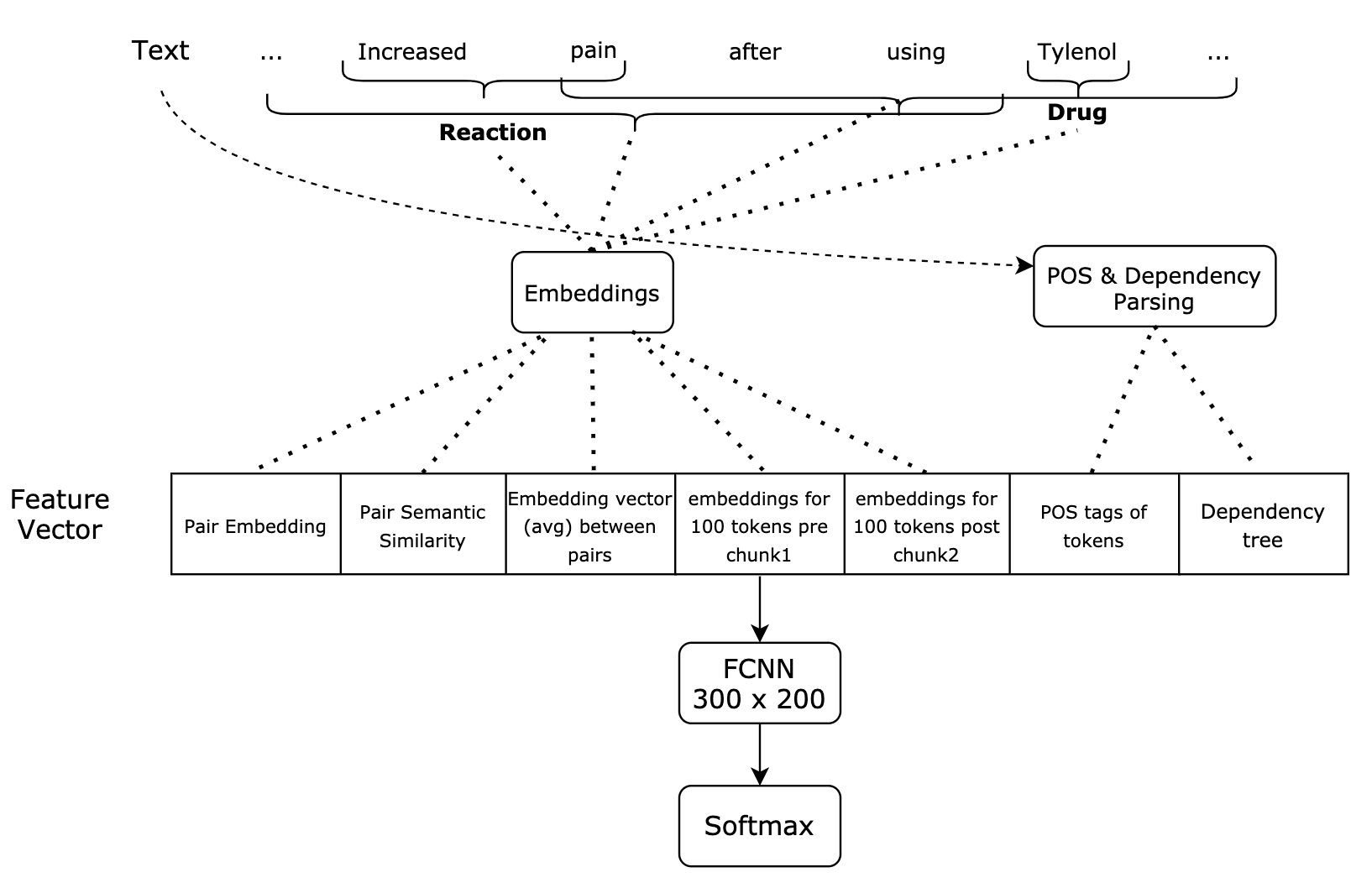}
  \caption{Overview of the first RE model. All the features are vertically stacked in a single feature vector. The feature vector is kept dynamic with additional padding for compatibility across different embedding sizes, and complex dependency structures.}
  \label{fig:re_model}
\end{figure}

For our first RE solution we rely on entity spans and types identified by the NER model to develop distinct features to feed to an FCNN for classification. At first we generate distinct pairs of entities (e.g. symptom-treatment), and then generate custom features for each pair. These features include semantic similarity of the entities, syntactic distance of the two entities, dependency structure of the entire document, embedding vectors of the entity spans, as well as embedding vectors for 100 tokens within the vicinity of each entity. Figure \ref{fig:re_model} explains our model architecture in detail. We then concatenate these features and feed them to fully connected layers with leaky relu activation. We also use batch normalisation after each affine transformation before feeding to the final softmax layer with cross-entropy loss function. We use softmax cross-entropy instead of binary cross-entropy loss to keep the architecture flexible for scaling on datasets having multiple relation types.

Our second solution focuses on a higher accuracy, as well as exploration of relations across long documents, and is based on \cite{DBLP:journals/corr/abs-1906-03158}. In our implementation, we implement the model in Apache Spark for scalability, take checkpoints from the BioBERT model, and train an end-to-end BERT model for RE. Similar to the first solution, this architecture also depends on the entity spans identified by the base NER model, and uses the entire document as context string while training the model. The original paper used sequence length of 128 tokens for the context string, which we keep constant, and instead experiment with the content of the context string, training data augmentation, and fine-tuning techniques.

We use Spark NLP's \cite{KOCAMAN2021100058} NER models \cite{DBLP:journals/corr/abs-2012-04005} as foundation for the RE models as these NER models provide entity spans required for performing RE. In a single inference pipeline, the RE models are placed sequentially after the the NER model, and are fed the results of the NER model, the context, embeddings, and dependency tree for feature generation. Apart from feature generation, the dependency tree also helps regularize candidate entity pairs for RE classification as we can eliminate pairs having a larger syntactic distance. This modular approach of arranging components reduces coupling and achieves a higher degree of memory and computational efficiency as components like sentences, tokens, and embeddings are shared between NER and RE models and don't need to be executed again. Since the NER model is essentially a token classifier and produces prediction per token, we convert the tokens to chunks using BIO tags.

\section{Experiments}

We test the models on public datasets, report evaluation metrics, and analyse the results on examples. In addition to public datasets, we explain the process of annotating and training models on new datasets. We then study the utility of applying RE for some use-cases like knowledge graph generation and improved entity resolution (the process of mapping entity chunks to medical codes).

\subsection{Performance on Public Datasets}
We test both model architectures on seven public datasets by using the official training-test split for training and testing the models, and report macro-average f1 scores for each one of them in Table \ref{tab:pub_metrics}. These datasets include the 2012 i2b2 challenge for evaluating temporal relations in clinical text \cite{article}, the 2010 i2b2/VA challenge on concepts, assertions, and relations in clinical text \cite{uzuner20112010}, the Drug-Drug-Interaction (DDI) dataset for linking drugs with dispositions and reactions \cite{HERREROZAZO2013914}, the Chemical–protein interaction (CPI) dataset for linking genes/proteins with drug chemicals \cite{Krallinger2017OverviewOT}, the Phenotype-Gene Relations (PGR) dataset for relating human phenotypes and genes \cite{sousa-etal-2019-silver}, the adverse drug events dataset for relating drugs with their reactions \cite{GURULINGAPPA2012885}, and the posology relations task based on the 2018 n2c2 task \cite{henry20202018}.  For the sake of brevity we don't delve into the details for each dataset, and specific details for each dataset can be found in the cited resources. As explained in Table \ref{tab:pub_metrics}, the BERT model achieves new SOTA metrics on 5 public datasets, and out performs the lighter FCNN model due to better contextual awareness. However, it is more than 3 time slower and has much higher memory requirements. Table \ref{tab:speed_comp} compares the speed difference of the two architectures. Hyperparameter setting and sample Python code for training an RE model from scratch is in Appendix A \& C.

\begin{table}[ht]
\centering
\begin{tabular}{ lccc }
\toprule
\textbf{Dataset} & \textbf{FCNN} & \textbf{BioBERT} & \textbf{Curr-SOTA} \\
\midrule
\textbf{i2b2-Temporal} & 68.7 & \textbf{73.6} & 72.41\\
\textbf{i2b2-Clinical} & 60.4 & \textbf{69.1} & 67.97\\
\textbf{DDI} & 69.2 & 72.1 & \textbf{84.1}\\
\textbf{CPI} & 65.8 & 74.3 & \textbf{88.9}\\
\textbf{PGR} & 81.2 & \textbf{87.9} & 79.4\\
\textbf{ADE Corpus} & 89.2 & \textbf{90.0} & 83.7\\
\textbf{Posology} & 87.8 & \textbf{96.7} & 96.1\\
\bottomrule
\end{tabular}
\caption{Macro-averaged F1 scores of both RE models on public datasets. FCNN refers to the Speed-Optimized FCNN architecture, while BioBERT refers to the Accuracy-Optimized BioBERT architecture. The SOTA metrics are obtained from \cite{DBLP:journals/corr/abs-2004-06216}, \cite{DBLP:journals/corr/abs-1903-09941}, \cite{10.1093/bioinformatics/btaa907}, \cite{DBLP:journals/corr/abs-2106-03598}, \cite{DBLP:journals/corr/abs-2001-07139}, \cite{DBLP:journals/corr/abs-2002-06424}, and \cite{DBLP:journals/corr/abs-2107-08957} respectively.}
\label{tab:pub_metrics}
\end{table}

\begin{table}[ht]
\centering
\begin{tabular}{ lcc}
\toprule
\textbf{Dataset} & \textbf{FCNN RE Model} & \textbf{BERT RE Model} \\
\midrule
\textbf{1k Notes} & 104 & 584\\
\textbf{10k Notes} & 925 & 5197\\
\bottomrule
\end{tabular}
\caption{Time in seconds taken by both models to run a sample batch of documents on a single linux machine with with 64GB of RAM, and an 8-Core CPU (without GPU hardware acceleration).}
\label{tab:speed_comp}
\end{table}

\subsection{Performance on Private Datasets}

In addition to the public datasets, we sampled approximately 5000 clinical notes and manually annotated them with the help of domain experts on the following guidelines: We selected general entities (e.g, body part, date, test result) that can compliment core entities (e.g, symptom, procedure, test) as the first entity and disjoint entity types - meaning the entities should not have relation among themselves - from the the core entities for the second entity as explained in Table \ref{tab:relation_pairs}. Since the first entity can relate to multiple entities in the second column, we can define the relation between the two entity types as one-to-many, and can keep the relation types to a minimum i.e. are the two entities related or not. This approach helps reduce annotation complexity resulting in faster annotation times, and a higher inter-annotator agreement. For annotation purposes we utilized the publicly available Annotation Lab tool.

\begin{table}[ht]
\centering
\begin{tabular}{ lcc}
\toprule
\textbf{model} & \textbf{Entity 1} & \textbf{Entity 2} \\
\midrule

\multirow{2}{*}{\textbf{re\_bodypart\_procedure\_test}} & \multirow{2}{*}{Body Part} & Procedure\\
& & Test\\
\midrule
\multirow{2}{*}{\textbf{re\_test\_result\_date}} & \multirow{2}{*}{Test} & Test Result\\
& & Date\\
\midrule
\textbf{re\_bodypart\_problem} & Body Part & Symptom\\
\midrule
\textbf{re\_test\_problem\_finding} & Test & Symptom\\
\midrule
\textbf{re\_bodypart\_directions} & Body Part & Direction\\

\bottomrule
\end{tabular}
\caption{Entity types in column \textit{Entity 2} do not have relations among themselves}
\label{tab:relation_pairs}
\end{table}





\section{Practical Applications of Relation Extraction}

\begin{figure}[h!]
  \includegraphics[width=0.45\textwidth]{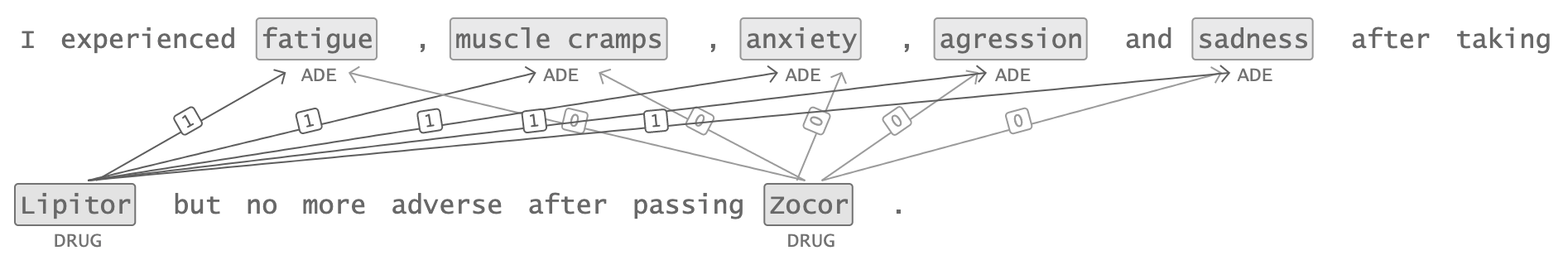}
  \caption{Output of the ADE RE model on sample data. Arrows with \textit{0} represent the two entities are not related, while \textit{1} represents that the reaction is caused by the drug.}
  \label{fig:ade_example}
\end{figure}

\begin{figure}[h!]
  \includegraphics[width=0.45\textwidth]{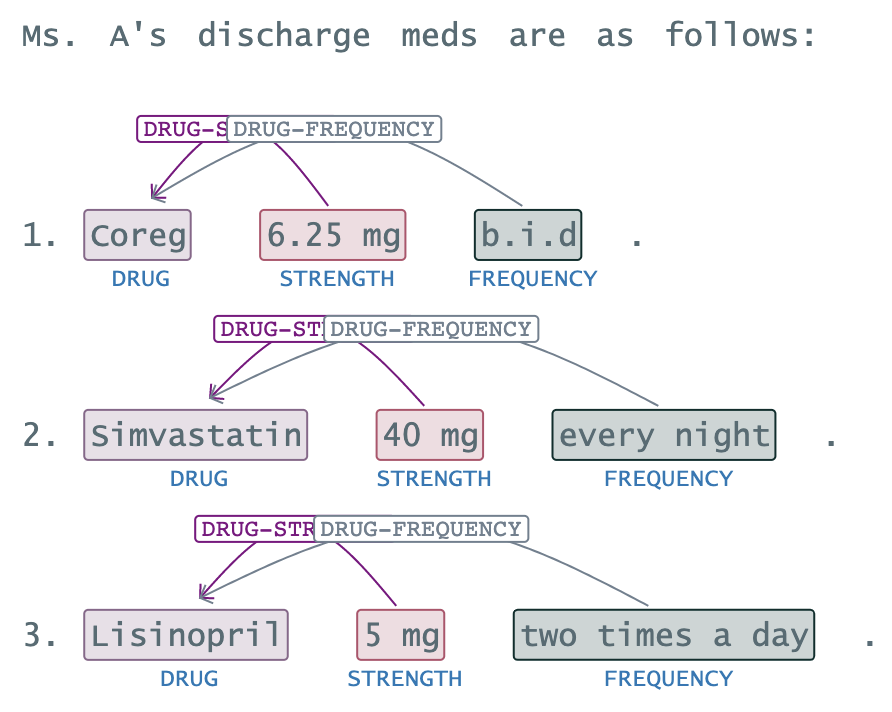}
  \caption{Output of the Posology RE model on sample data.}
  \label{fig:posology_example}
\end{figure}

The ability to semantically relate entities paves way for a lot of opportunities and use-cases. For example, the RE model for Adverse Drug Events can be used to identify drugs that caused reactions in large trial datasets. Figure \ref{fig:ade_example} shows the output of running the ADE RE model on sample text. Similarly, lab results, discharge notes, and prescriptions can be parsed to get a structured output as illustrated in Figure \ref{fig:posology_example}.

In addition of using the public models, following are some of the use-cases we explored with our general-purpose models:

\subsection{Generating Knowledge Graph with Relations}
\begin{figure}[h!]
  \includegraphics[width=0.45\textwidth]{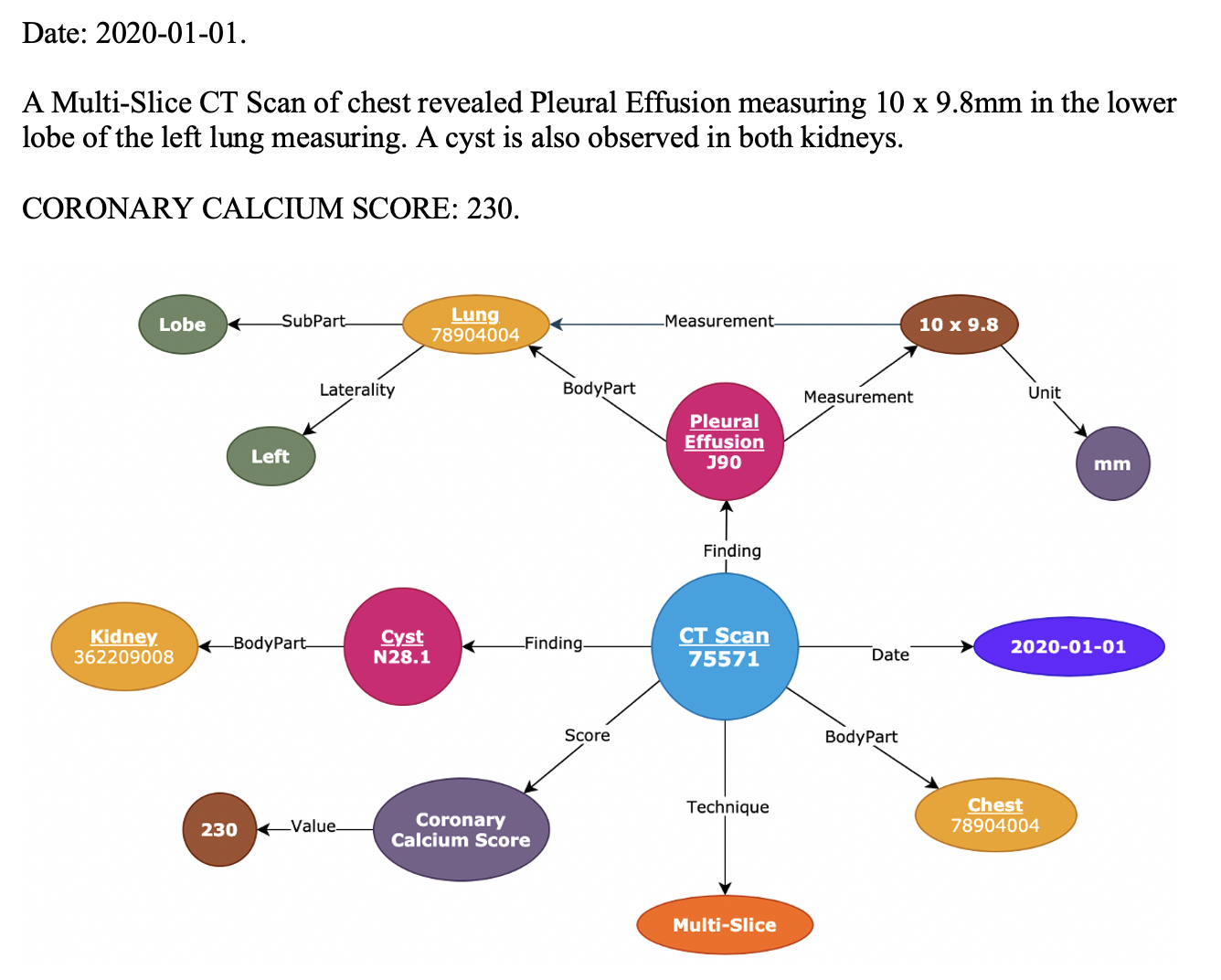}
  \caption{A graphical representation (with CPT, ICD \& SNOMED codes) of the structured data extracted from a sample text.}
  \label{fig:re_graph}
\end{figure}

Most notable benefit of RE is the ability to generate knowledge graphs from unstructured text. For this experiment, we used pretrained Spark NLP NER models and the general-purpose RE models explained in the previous section to process medical reports with the primary goal of generating a concise structured output of a report. For instance, we relate procedures with dates and findings to recognize dates of a procedure and its findings along with any existing condition. We use the relations between body parts and procedures to get more specific details of the location of the procedure. Similarly, relating body parts with findings like test results and measurements can add more details to the final output in specific use-cases. More granularity can be achieved by having further subdivisions of body parts. For instance, in our experiment, we divide the body part in three parts; the primary body part (e.g, lung), a sub-part (e.g, lobe), and direction/laterlity (e.g, left) of the body part. In practice, these specifc entities trickle from the NER model down to the RE models. A graph generated from a sample report can be seen in Figure \ref{fig:re_graph}.

\begin{figure}[h!]
  \includegraphics[width=0.45\textwidth]{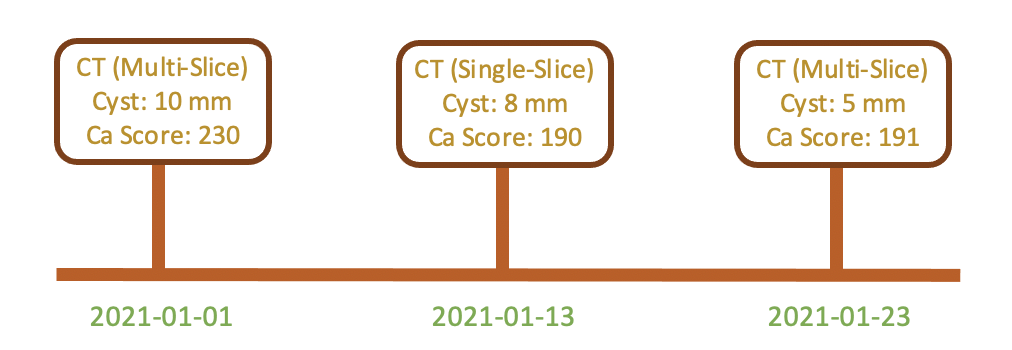}
  \caption{A sample timeline of a patient showing calcium score trend, and evolution of cyst over multiple scans in a month.}
  \label{fig:pat_timleine}
\end{figure}

Furthermore, the structured data can help create a patient timeline which can show progress of a certain condition over a certain duration. A sample timeline monitoring coronary calcium score and cyst can be seen in Figure \ref{fig:pat_timleine}. Such information can be used to analyse multiple trends like effectiveness of a drug for treating a certain condition on large datasets.
 
\subsection{Enriching Chunks for more Accurate Coding}

Entity Resolver models map entity chunks to medical codes like CPT \cite{cpt}, ICD \cite{icd}, SNOMED \cite{snomed}, MeSH \cite{mesh}, RxNorm \cite{rxnorm} etc based on semantic similarity. This task becomes challenging due to two major reasons. First, the inherent noise of the text like abbreviations, acronyms, and synonyms can result in false positive results. Second, medical codes are sensitive to variables like severity, location in human body, administration type, diagnosis method, etc; For a given condition or treatment, there could be different codes (within the same ontology) depending on the aforementioned factors. This challenge is more prominent in ontologies with wider vocabularies like SNOMED.

RE provides solution to both problems; First, it intrinsically cleans the input for the resolver models of stop words and noise without additional effort. Second, it adds additional information to the core entity chunks from surrounding context; With the help of relations, simple entities can be enriched with precise information to get accurate codes. For example, a chunk \textit{CT Scan} - identified as a procedure - can be enriched with the imaging technique to achieve a more accurate CPT/SNOMED code. Enriching it further with the location of the procedure (e.g, chest) would result in an even accurate chunk that can be resolved to a more specific CPT/SNOMED code. Table \ref{tab:resolver_comp} compares base chunks with enriched chunks that include body parts, demonstrating the benefits of enriched entity chunks for improved coding.

\begin{table*}
\centering
\begin{tabular}{ lcccc}
\toprule
\textbf{Ontology} & \textbf{Base Chunk} & \textbf{Base Code} & \textbf{Enriched Chunk} & \textbf{Enriched Code}\\
\midrule
CPT & CT Scan & 3324F & CT Scan Chest (multi-slice) contrast  & 71260 \\
SNOMED CT & CT Scan & 169072007 & CT Scan Chest (multi-slice) contrast  & 169069000 \\
ICD-10 & Lesion & L98. 9   & Lesion liver & K76. 9\\
SNOMED & Lesion & 300577008   & Lesion liver & 300331000\\
\bottomrule
\end{tabular}
\caption{Comparison of entity resolution results - more enriched and specifc entity chunks result in a more accurate code.}
\label{tab:resolver_comp}
\end{table*}

\section{Conclusion}

In this paper we presented two new model architectures for RE while enabling scalability. We then tested the models on public datasets and reported evaluation metrics. The model metrics show that the BioBERT based model outperforms the lighter FCNN model, and obtains new state-of-the-art accuracy on on three benchmarks. However, for datasets with a small number of relation types, the simpler FCNN model may be a compelling option not only due to faster run times, but also much lower memory requirements compared to the BioBERT model, allowing to process larger datasets on commodity hardware. We also explain how to train RE models from scratch and describe the design behind the pre-trained models available as part of Spark NLP library.

We then study practical use cases where RE plays the salient role of linking entities together to generate knowledge graphs, patient timelines, and structured summaries of medical notes. Relating dates to primary procedures and problems can help create a timeline for each patient. Finally, using granular NER models together with discrete RE models to clean and enrich entity chunks enables better entity resolution to clinical codes.

Given the complex nature of RE, and the pivotal role of contextual information, a common approach is to limit relations within a certain syntactic span as even BERT models have token sequence limit. A future research direction could be to focus on improving contextual representation of large documents to allow relations over lengthy contextual spans. A second future research direction is to test whether auxiliary data - either from medically annotated data or through transfer learning from healthcare-specific language models - can deliver higher accuracy Relation Extraction on the same neural network architectures.


\section{Appendices}

\appendix
\section{A. Hyperparameter Settings}

Since optimal hyperparameter values vary for each dataset, a range of values which performed best in all the datasets can be seen in Table \ref{tab:re_params}.
\begin{table}[ht]
\centering
\begin{tabular}{ lccc}
\toprule
\textbf{Parameter} & \textbf{FCNN} & \textbf{BERT} & \textbf{Tested Range} \\
\midrule
\textbf{Dropout rate} & 0.5 & 0.2 & 0.2-0.5\\
\textbf{Batch size} & 64 - 128 & 64 & 8-128\\
\textbf{Learning rate} & 0.0003 & 0.001 & 0.0001-0.005\\
\textbf{Epoch} & 50 - 70 & 2-5 & 1-100\\
\textbf{Optimizer} & Adam & Adam & Adam\\
\textbf{LR Decay} & 0.005 & 0.005 & 0.002-0.005\\
\bottomrule
\end{tabular}
\caption{Best performing hyperparameter value ranges on multiple datasets.}
\label{tab:re_params}
\end{table}

\section{B. Preparing training data for RE model in Spark NLP}
Since RE is a classification task, the primary inputs are the context string (sentence), and a pair of entities. If there are multiple pairs in a single context string, we treat them as disjoint inputs as each input encapsulates the required inputs like entity chunk pairs and context - which are then used to create input features. We can create a csv formatted file where each row is a training example for the model, and contains the aforementioned inputs. Exact schema of the training file can be found in the training notebook \cite{re_training_code}.

\section{C. Training an RE Model in Spark NLP}
Code for training an RE mode is provided as a google colab notebook \cite{re_training_code}. As majority of the public datasets are protected and can not be shared, they need to be obtain from their official websites and converted to the required format before training.

\newpage
\bibliography{main}

 
\end{document}